\documentclass[journal]{IEEEtran}
\usepackage{amsfonts}
\usepackage{amssymb,amsmath,color,graphicx}
\usepackage{subfigure}
\usepackage{cite}
\usepackage{comment}
\usepackage{verbatim}
\usepackage{algorithm}
\usepackage{algorithmic}
\usepackage{xcolor}
\usepackage{placeins}
\usepackage{color,soul}
\usepackage{setspace}
\usepackage{float}
\usepackage{bm}
\usepackage{dsfont}
\usepackage{xr}
\usepackage{tabu}
\usepackage{tabularx}
\usepackage{longtable}
\usepackage[utf8]{inputenc}
\usepackage{pifont}
\usepackage{wasysym}
\usepackage{lipsum}
\usepackage{booktabs}
\usepackage{tabulary}
\usepackage{textcomp}
\usepackage{booktabs}
\usepackage{multirow}
\usepackage{multicol}
\usepackage{wrapfig}
\usepackage{mathtools}
\usepackage{subfig}
\usepackage{booktabs}
\usepackage{array, makecell}
\usepackage{caption}
\ifCLASSINFOpdf
\else
\fi

\usepackage{array}

\begin{document}
%
\title{Gradient-Enhanced Physics-Informed Neural Networks for Power Systems Operational Support}
%

\author{Mostafa~Mohammadian,~\IEEEmembership{Student~Member,~IEEE,}
        Kyri~Baker,~\IEEEmembership{Member,~IEEE,}
        and~Ferdinando~Fioretto
\thanks{M. Mohammadian and K. Baker are with the College of Engineering and Applied Science, University of Colorado Boulder, Boulder, CO $80309$ USA. (e-mail:mostafa.mohammadian@colorado.edu; kyri.baker@colorado.edu).}
\thanks{F. Fioretto is affiliated with the Electrical
Engineering and Computer Science Department, Syracuse University, Syracuse, NY $13244$, USA. (e-mail: ffiorett@syr.edu)}
\thanks{This work was supported by the National Science Foundation CAREER award 2041835. Additionally, this work utilized the Summit supercomputer, which is supported by the National Science Foundation (awards ACI-$1532235$ and ACI-$1532236$), the University of Colorado Boulder, and Colorado State University. The Summit supercomputer is a joint effort of the University of Colorado Boulder and Colorado State University.}
}

\maketitle

\begin{abstract}
The application of deep learning methods to speed up the resolution of challenging power flow problems has recently shown very encouraging results. However, power system dynamics are not snap-shot, steady-state operations. These dynamics must be considered to ensure that the optimal solutions provided by these models adhere to practical dynamical constraints, avoiding frequency fluctuations and grid instabilities. 
Unfortunately, dynamic system models based on ordinary or partial differential equations are frequently unsuitable for direct application in control or state estimates due to their high computational costs. 
To address these challenges, this paper introduces a machine learning method to approximate the behavior of power systems dynamics in near real time.
The proposed framework is based on gradient-enhanced physics-informed neural networks (gPINNs) and encodes the underlying physical laws governing power systems. 
A key characteristic of the proposed gPINN is its ability to train without the need of generating expensive training data. The paper illustrates the potential of the proposed approach in both forward and inverse problems in a single-machine infinite bus system for predicting rotor angles and frequency, and uncertain parameters such as inertia and damping to showcase its potential for a range of power systems applications.
\end{abstract}


%
\IEEEpeerreviewmaketitle

\section{Introduction}
\IEEEPARstart{I}n recent years, with the dramatic growth of available data and computing resources, there has been a revolution in the successful application of Artificial Neural Networks (ANN), also commonly referred to Deep Neural Networks (DNN) and Deep Learning (DL), in power systems. DNNs can tackle a range of highly complex tasks and are especially useful when a precise mathematical representation of the problem cannot be acquired, or when conventional techniques are too computationally or numerically challenging. The applications of ANNs and DNNs span many scientific disciplines, such as image recognition, speech recognition and translation, cognitive science, to name a few examples \cite{Zhang9039675}. In power systems, some recent works have been focused to solve computational problems both in dynamics and optimization at a fraction of the time required by traditional approaches due to the harnessing the benefit of historical data and moving most of the computational burden to an off-line setting \cite{Zamzam_learn_19, fioretto2020predicting, Mostafa21, Biagioni22}.

At first glance, it seems very challenging to train a deep learning algorithm (model) to reliably determine a nonlinear map from a few, potentially very high-dimensional, input and output data pairs. However, there is a great amount of prior knowledge in many cases relevant to the modeling of physical systems, e.g. power systems, that is currently not being exploited in modern machine learning techniques. Hence, machine learning approaches used in power systems (and other physical systems) were, however, mainly agnostic to the underlying physical model until recently. This made them extremely reliant on the quality of the training data, necessitating enormous training datasets and, in some cases, complicated neural network structures. However, in the course of evaluating complicated systems or problems, the expense of data collecting is frequently prohibitive, and we are forced to draw conclusions and make decisions based on partial information. On the other hand, it is widely acknowledged that current state-of-the-art machine learning tools (e.g., deep/recurrent neural networks), which are also often applied in power systems research, lack robustness and fail to provide any guarantees of convergence when used in the small data regime, i.e., when only a few training instances are available \cite{RAISSI2019686}. 

Despite recent efforts to make dataset construction more effective, which have yielded promising results \cite{Venzke2019, Thams8600355, OPFLearn_ISGT}, creating the needed training dataset size still necessitates a significant amount of computational effort. In this paper, we lessen the reliance on training data and sophisticated neural network topologies by utilizing the underlying physical rules provided by power system models during training of the neural network. To achieve this, we propose the use of physics-informed neural networks (PINNs), which can directly incorporate power systems dynamics into the training of these neural networks. This has been initially studied in \cite{Misyris2019} in the context of the dynamical generator swing equations, which we also consider here. However, departing from this work, we consider more advanced PINN architectures - specifically, a gradient-enhanced PINN (gPINN) - as well as test these models under a wider variety of system conditions. We lastly propose a method to generalize the models to different boundary conditions using transfer learning, which greatly improves the computational efficiency. 


The use of PINNs and gPINNs allows these deep learning models to directly incorporate dynamic relationships. This is performed by embedding the power system differential and algebraic equations into the training procedure (loss of the neural network) using automatic differentiation (AD) \cite{BaydinDifferentiation}. Exploiting advances in AD, which have been widely integrated into many deep learning frameworks such as
Tensorflow \cite{AbadiABBCCCDDDG16} and PyTorch \cite{Paszke2017}, we can directly compute derivatives of neural network outputs during training, such as the rotor angle, and develop neural networks that accurately capture the rotor angle and frequency dynamics.
Additionally, the basic formulation of the gPINN/PINN training (forward problem) does not require labeled data, e.g., results from other simulations or experimental data. Unsupervised gPINNs/PINN only require the evaluation of the residual function, which alleviates the problem of label data creation \cite{10.3389/fdata.2021.669097}. Note however that providing simulation or experimental data for training the network in a supervised manner is also possible and necessary for inverse problems \cite{Mishra2020}.
Our proposed method requires less initial training data,  can result in smaller neural networks, and achieves good performance under a variety of different system conditions.

Gradient-enhanced physics-informed neural networks (PINN) are a novel technology that could lead to a new class of numerical solvers \cite{Yu2021, RAISSI2019686} and dynamic state estimation approaches \cite{Zhao8624411}. Within power systems, they have the ability to solve systems of differential-algebraic equations in a fraction of the time required by traditional methods, to directly determine the value of state variables at any time instant $t_1$ (without the need to integrate from $t_0$ to $t_1$), and to solve higher-order differential equations without the need to introduce additional variables to solve a first-order system.

In this paper, we focus on power system dynamics and use the swing equation as an example to demonstrate the main ideas for the application of gradient-enhanced physics informed neural networks of \cite{Yu2021} in power systems. Such cases are abundant in the study of power systems, where longstanding developments of mathematical physics have shed tremendous insight on how such systems are structured, interact, and dynamically evolve in time. Moreover, we show how the same methods may be used to estimate unknown or uncertain parameters like inertia and damping, in addition to solving ordinary differential equations. Note that the main purpose of utilizing our proposed method is to \emph{assist} human dispatchers rather than directly replacing current controllers. Using these models can help operators simulate a wide variety of dynamic system conditions very quickly, helping them make informed decisions during operation with the aim of increasing security and reducing costs. Hence, The outcome of the proposed model can be used for introducing operation-oriented preventive measures in which dispatchers could use to avoid critical situations.

The main contributions of this paper center around demonstrating the efficacy of PINN models in both finding solutions to the swing equation and estimating unknown parameters. While this study initially focuses on a single-machine, infinite bus (SMIB) system as in \cite{Misyris2019}, we simulate much more dynamic conditions using different PINN architectures, which is an important next step towards using deep learning in the broader field of power system dynamics. Thus, the main contributions of this work are as follows:

\begin{itemize}
    \item  We develop a model which uses deep learning to reliably calculate solutions to the swing equation, as a result, can rapidly obtain values such as as load angle and frequency in the swing equation;
    \item The model is capable of accurately estimating uncertain power system parameters from limited measurements, which becomes increasingly important as generators age and parameters naturally change; 
    \item Lastly, the model can quickly generalize to different initial conditions through transfer learning, meaning that operators can use the model as an advisory tool to simulate a wide variety of possible system states. 
\end{itemize}

The remainder of this paper is organized as follows. In Section \ref{sec: methods}, after introducing the physics-informed neural networks architecture, we present the extension to g-PINN. Section \ref{sec: model} briefly describes the employed power system model and gPINN used for the prediction task. In Section \ref{sec: result}, we present the simulation results demonstrating the performance of physics-informed neural networks and gPINN methods. Finally, conclusions and future work are given in Section \ref{sec: conclude}.

\section{Methods} \label{sec: methods}

In this section, we first provide an overview of physics-informed neural networks (PINNs) and then present the method of gradient-enhanced PINNs to improve the accuracy and training efficiency of PINNs in our given context. The PINNs aim at approximating the solution of a system of one or more differential, possibly non-linear equations, by explicitly encoding the differential equation formulation in the neural network. Without loss of generality, we consider the following parametrized and nonlinear PDE of the general form:
\begin{align}
\begin{split}
    &\bm{u}(\bm{x}, t)_t + \mathcal{N}_x[\bm{u}(\bm{x}, t), \lambda] = 0, \qquad \bm{x} \in \Omega, t \in [0, T]  \label{eq: PDE}\\ 
    &\mathcal{B}(\bm{u}(\bm{x}, t)) = 0, \qquad (\bm{x}, t) \in  \partial \Omega
\end{split}
\end{align}
where $t$ and $\bm{x}$ represent time and space coordinates, $\bm{u}(t, \bm{x})$ denotes the latent (hidden) solution, $\mathcal{N}_x[\cdot; \lambda]$ is a nonlinear differential operator connecting the state variables $\bm{u}$ with the system parameters $\lambda$, operator $\mathcal{B}$ denotes the boundary condition of a partial differential equation, subscripts denote partial differentiation, $\Omega$ is a subset of $\mathbb{R}^D$, $\partial \Omega$ is the boundary of $\Omega$, and $[0, T]$ is the time interval within which the system evolves. We note that the initial condition can be  treated as a special type of Dirichlet boundary condition on the spatio-temporal domain. Moreover, the model parameters $\lambda$ can be unknown or fixed. In case $\lambda$ is unknown, the problem of approximating the solution of function (\ref{eq: PDE}) would be a problem of system identification, where we seek parameters $\lambda$ for which the expression in (\ref{eq: PDE}) is satisfied. It is noteworthy  that PINNs can also be designed to effectively integrate Ordinary Differential Equations (ODE) \cite{Wang2020}.

In the next step, we need to restrict the neural network $\hat{\bm{u}}(\bm{x}, t)$ to conform to the physics imposed by the PDE and boundary conditions. In practice, we restrict  $\hat{\bm{u}}$ on some scattered points (e.g., clustered points, or randomly distributed points  in the domain), i.e., the training data $\mathcal{T} = \{(t_1, \bm{x}_1), (t_1, \bm{x}_1),  \ldots, (t_{|\mathcal{T}|}, \bm{x}_{|\mathcal{T}|})\}$ of size $|\mathcal{T}|$. Moreover, $\mathcal{T}$ consists of two sets $\mathcal{T}_f \subset \Omega$  and $\mathcal{T}_b \subset \partial\Omega$, which are the points in the domain and on the boundary, respectively. Following the original work of \cite{RAISSI2019686} for PINNs, we then proceed by approximating $\bm{u}(\bm{x}, t)$ by a deep neural network, and define $f_{\bm{\theta}}(\bm{x}, t)$ (resulting physics-informed neural network parameterized by $\bm{\theta}$) to be given by the left-hand-side of  (\ref{eq: PDE}); i.e.,
\begin{align}\label{eq: f_theta}
    f_{\bm{\theta}}(\bm{x}, t) \coloneqq \frac{\partial}{\partial t}\bm{u}(\bm{x}, t) + \mathcal{N}_x[\bm{u}(\bm{x}, t), \lambda].
\end{align}
Note that if the system parameters $\lambda$ are known in advance, the nonlinear operator $\mathcal{N}_x[\bm{u}, \lambda]$ reduces to $\mathcal{N}_x[\bm{u}]$.

\begin{figure}[t]
\centering
\includegraphics[scale=0.9, trim={0.25cm 2cm 7.9cm 1cm},clip]{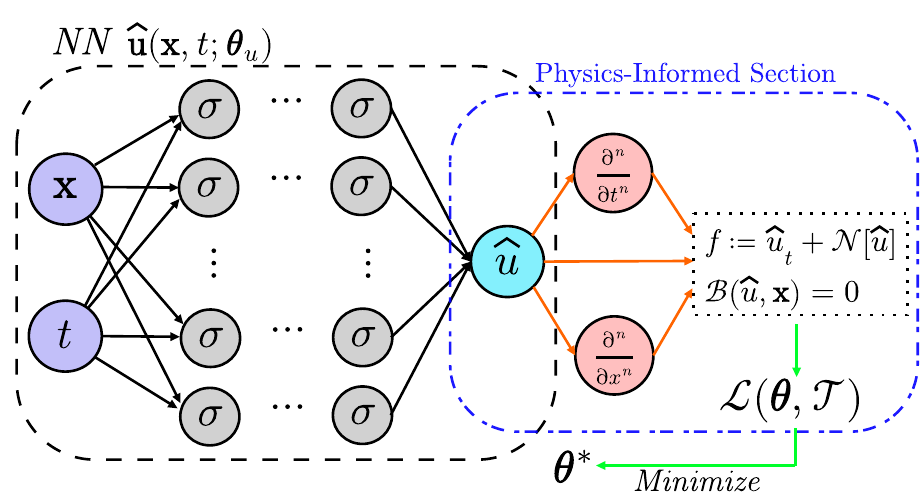}
\caption{Schematic of a typical PINN.}
\vspace{-5mm}
\label{fig: schematic of PINN}
\end{figure}

Fig. \ref{fig: schematic of PINN} shows the general architecture of a sample PINN. In a PINN, we first construct a neural network $\hat{\bm{u}}(\bm{x}, t; \bm{\theta})$ with the trainable parameters $\bm{\theta}$ as a surrogate or approximation of the solution $\bm{u}(\bm{x}, t)$, which takes inputs $(\bm{x}, t)$, called collocation points, in the simulation domain and outputs a vector with the same dimension as $\bm{u}$. Here, $\bm{\theta} = \{\bm{W}^l, \bm{b}^l\}_{1\leq l \leq L}$ is the set of all weights and biases in the neural network $\hat{\bm{u}}(\bm{x}, t)$, called the approximator/surrogate network. One advantage of choosing neural networks as the surrogate of $\bm{u}$ is that we can take the derivatives of $\hat{\bm{u}}$ with respect to time $t$ and system inputs $\bm{x}$ by applying the chain rule for differentiating compositions of functions using the automatic differentiation (AD) of the same neural network, which is conveniently integrated in machine learning packages such as TensorFlow and PyTorch. As a result, the neural network predicting $f(\bm{x}, t)$ has the same parameters as the network representing $\bm{u}(\bm{x}, t)$, albeit with different activation functions.  The shared parameters between the neural networks $\bm{u}(\bm{x}, t)$ and $f(\bm{x}, t)$ are optimized by minimizing the mean squared error loss

\begin{align}
	\mathcal{L}(\bm{\theta}; \mathcal{T}) = w_f\mathcal{L}_f(\bm{\theta}; \mathcal{T}_f) + w_b\mathcal{L}_b(\bm{\theta}; \mathcal{T}_b)\label{eq: PINN-loss}, 
\end{align}where
\begin{align*}
\begin{split}
    &\mathcal{L}_f(\bm{\theta} ; \mathcal{T}_f) = \frac{1}{|\mathcal{T}_f|}\sum_{(\bm{x}, t) \in \mathcal{T}_f} |f(\bm{x}, t)|^2\\ 
    &\mathcal{L}_b(\bm{\theta}; \mathcal{T}_b) =\frac{1}{|\mathcal{T}_b|}\sum_{(\bm{x}, t) \in \mathcal{T}_b} |\mathcal{B}( \hat{\bm{u}}, \bm{x})|^2
\end{split}
\end{align*}
and $w_f$ and $w_b$ are the weights. Here, to measure the discrepancy between the neural network  $\hat{\bm{u}}$ and the constraints imposed by the PDE in (\ref{eq: PDE}), we consider the loss function defined as the weighted summation of the $L^2$ norm of residuals for the equation and boundary conditions. Specifically, the loss $\mathcal{L}_f$ enforces the physics of the dynamical system by (\ref{eq: PDE}) at a finite set of collocation points and $\mathcal{L}_b$ corresponds to the boundary and initial conditions. Given a training data set and known system parameters $\lambda$, in the last step, we seek to find the parameters (weights and biases) of the neural networks by minimizing the loss $\mathcal{L}(\bm{\theta}; \mathcal{T})$.

One key advantage of PINNs is that the same formulation can be utilized not only for forward problems but also for inverse PDE-based problems. If the parameter in (\ref{eq: PDE}) is unknown, and instead we have some additional $\bm{u}$ measurements on the set of points $\mathcal{T}_i$, then, as shown in \cite{Lu2019}, we add an extra data loss
\begin{align}
    \mathcal{L}_i(\bm{\theta}, \bm{\lambda} ; \mathcal{T}_i) = \frac{1}{|\mathcal{T}_i|}\sum_{(\bm{x}, t) \in \mathcal{T}_i} |\hat{\bm{u}}(\bm{x}, t)- \bm{u}(\bm{x}, t)|^2. 
\end{align}
To simultaneously learn the unknown parameters with the solution $\bm{u}$, then our new loss function is  described as
\begin{align}\label{eq: tot-PINN-loss}
	\mathcal{L}(\bm{\theta}, \bm{\lambda}; \mathcal{T}) = w_f\mathcal{L}_f(\bm{\theta}, \bm{\lambda}; \mathcal{T}_f) &+ w_b\mathcal{L}_b(\bm{\theta}, \bm{\lambda}; \mathcal{T}_b)\\
	&+ w_i\mathcal{L}_i(\bm{\theta}, \bm{\lambda} ; \mathcal{T}_i)\nonumber. 
\end{align}    


In PINNs, the PDE residual $f$ is only enforced to be zero; since $f_\theta(\bm{x}, t)$ is zero for any $(\bm{x}, t)$, we know that the derivatives of $f$ are also zero in the simulation domain. Here, the gradient-enhanced PINNs is proposed in \cite{Yu2021} to enforce the derivatives of the PDE residual to be zero as well, i.e.,

\begin{align}
	\nabla f(\bm{x}, t) = (\frac{\partial f}{\partial x_1}, \frac{\partial f}{\partial x_2}, \cdots, \frac{\partial f}{\partial x_D}, \frac{\partial f}{\partial t}),  \bm{x} \in \Omega, t \in [0, T]
\end{align}
Then the loss function of gPINN would be:
\begin{align}
	\mathcal{L} = w_f\mathcal{L}_f + w_b\mathcal{L}_b + w_i\mathcal{L}_i + \sum_{i=1}^{D} w_{g_i} \mathcal{L}_{g_i}(\bm{\theta}; \mathcal{T}_{g_i})\label{eq: tot_gPINN-loss}, 
\end{align}
where the loss of the derivative $\mathcal{L}_{g_i}(\bm{\theta}; \mathcal{T}_{g_i})$ with respect to $x_i$ is
\begin{align*}
	\mathcal{L}_{g_i}(\bm{\theta}; \mathcal{T}_{g_i}) = \frac{1}{|\mathcal{T}_{g_i}|}\sum_{(\bm{x}, t) \in \mathcal{T}_{g_i}} 	|\frac{\partial f}{\partial x_i}|^2.
\end{align*}
Here, $\mathcal{T}_{g_i}$ is the set of residual points for the derivative $\frac{\partial f}{\partial x_i}$ which can be different from set $\mathcal{T}_f$. As we will show in our simulation results, by enforcing the gradient of the PDE residual function, the accuracy of the predicted solutions for $\bm{u}$ is improved by the gPINN and requires fewer training points.  Moreover, the gPINN improves the accuracy of the predicted solutions for $\frac{\partial \bm{u}}{\partial x_i}$ which is beneficial for our study.

\section{Physical Model for Power System Dynamics} \label{sec: model}

In three-phase, high-voltage power transmission systems, synchronous generators of the system accelerate or decelerate with respect to the synchronously rotating air gap magnetomotive force, to adapt to changing power transfer requirements that occur during system disturbances. In electrical power systems networks, the frequency varies constantly based on system dynamics. Modeling power system dynamics from oscillations and transients using time-synchronized measurements can provide real-time information, including angular displacements, voltage and current phasors, frequency changes, and rate of signal system decay from positive-sequence components that can help system operators or dispatchers in decision making. 

In general, the dynamics of a generator is defined by its internal voltage phasor. In the context of transient stability assessment, the internal voltage magnitude is usually considered to be constant due to its slow variation in comparison to its angle \cite{Vu2017}. Therefore, power system dynamics are described by the swing equation in their simplest and most common form, neglecting transmission losses and bus voltage deviations. As such, for each generator $k$, the resulting the swing equation that governs the dynamics of the synchronous generator can  be described by \cite{Misyris2019} 
\begin{align}\label{eq: swing}
	m_k\ddot{\delta}_k + d_k\dot{\delta}_k + P_{e_k} - P_{m_k} = 0,
\end{align}where $m_k > 0 $ is the dimensionless moment of inertia of the generator, $d_k > 0 $ represents primary frequency controller action on the governor (called damping coefficient), $P_{m_k}$ is the input shaft power producing the mechanical torque acting on the rotor of the generator,  $P_{e_k}$ is the electrical power output in per unit, ${\delta}_k$ denotes the voltage angles behind the transient reactance, and $\dot{\delta}_k$ is the angular frequency of the $k^{th}$ generator, often also represented as $\omega_k$.

\begin{figure}[t]
\centering
\includegraphics[scale=1.2, trim={0.28cm 4cm 10cm 1.9cm},clip]{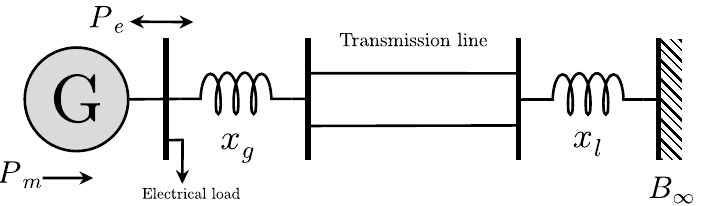}
\caption{Configuration of a single machine infinite bus system system.}
\vspace{-6mm}
\label{fig: SMIB}
\end{figure}

The single machine infinite-bus (SMIB) system, as depicted in Fig. \ref{fig: SMIB}, where $G$ is a synchronous generator and $B_{\infty}$ denotes a standard infinite bus, has been widely used to perceive and analyze the fundamental dynamic phenomena occurring in power systems. As the focus of this paper is on the introduction of gradient-enhanced physics-informed neural networks for power systems, we will use this system as a guiding example. We note though that our proposed framework is general and can be applicable to even more complex dynamical systems defined by the differential equations in the power systems. The swing equation (\ref{eq: swing}) for the SMIB system can be written as follows:
\begin{align}\label{eq: swing SMIB}
	m_g \, \ddot{\delta} + d_g \, \dot{\delta} + E_g E_{\infty} B_{g\infty} \sin(\delta) - P_{m} = 0,
\end{align}
where $\delta$ is the phase difference between the voltage vector of the generator and the infinite bus, $E_g$ is the voltage behind the transient reactance of the generator,  $E_{\infty}$ is the infinite bus voltage, and $B_{g\infty} \coloneqq \frac{1}{x_g+x_l}$ is the combined susceptance in between. we will show how gradient enhanced physics-informed neural networks can accurately estimate both rotor angle $\delta$ and angular frequency $\dot{\delta}$ (or $\omega$) of the swing equation (\ref{eq: swing SMIB}) at any time instant $t$ for a range of mechanical power $P_m$, and can identify uncertain parameters such as $m_g$ and $d_g$.

At the first step for solving the forward problem, it is assumed that the system parameters for a given synchronous generator $\lambda \coloneqq \{m_g, d_g \}$ are known priori and the voltages $E_g$ and $E_\infty$ are fixed and constant. As a result the system input to our proposed gPINN (and PINN) model is defined as $t$ (in our simulation domain). Despite conventional numerical solvers in simulating multi-physics problems, which necessitate the conversion of higher-order ordinary differential equations (ODEs) to first-order in order to solve them (by introducing additional variables), gPINNs can directly incorporate higher-order ODEs, as it will be shown in (\ref{eq: gPINNswing}). Incorporating (\ref{eq: swing SMIB}) to the neural network in Section \ref{sec: methods}, function (\ref{eq: f_theta}) is given
\begin{align}
		&u(t) \coloneqq \delta(t),\\ 
	&f_{\bm{\theta}}(t) = m_g \, \ddot{\delta} + d_g \, \dot{\delta} + E_g E_{\infty} B_{g\infty} \sin(\delta) - P_{m} = 0\label{eq: gPINNswing},\\
    &  \qquad P_m \in [P_{\min}, P_{\max}], t \in [0, T] \nonumber
\end{align}
The interval $[0, T]$ can be specified based on the time horizon of interest for the dynamic simulation. The domain $\Omega$ of the input $P_m$ is restricted to $[P_{\min}, P_{\max}]$ due to the capability of generator. The neural network output is only $\delta(t)$, rotor angular displacement with respect to the synchronous reference axis. After the training phase, the angular frequency signal $\omega \coloneqq \dot{\delta}$ is extracted as a function of the estimated angle $\delta$ using  the  automatic  differentiation (AD)  of  the  same  neural  network. As a result, the prediction error of the frequency $\omega$ depends on the prediction error of the angular position $\delta$ and the differential method.

In the second part, the system parameters $\lambda \coloneqq \{m_g, d_g \}$ are also the problem of interest to be determined by the gPINN model as well as angular position and frequency. For system operators to avoid substantial frequency deviations and preserve frequency stability, information about power system factors such as system inertia is critical. Varying power-infeed from converter-based generation units introduces great uncertainty on system parameters such as inertia and damping and, therefore, has to be estimated (or predicted) at regular time intervals \cite{Zografos8273824}. The problem of system identification and data-driven discovery of partial differential equations can be addressed with gradient-enhanced physics-informed neural networks. Hence, we define $m_g$ and $d_g$ as unknown parameters in (\ref{eq: gPINNswing}). Here, the topology of the gradient-enhanced physics-informed neural network stays unchanged, with the exception that when minimizing (\ref{eq: tot_gPINN-loss}) during neural network training, a subset of the system parameters are now handled as new variables.
\vspace{-5mm}
\section{Simulation Results} \label{sec: result}
In this section, we perform a series of tests to support the proposed methodology.  The goal of our experiments is to show that our gradient-enhanced physics-informed neural network is indeed capable of approximating the analytical solution given in (\ref{eq: gPINNswing}). The neural network models (PINNs and gPINNs) are constructed based on the given initial conditions. In this paper, the neural network approximation results are compared to the true solutions in order to test the models. The experimental design and findings of the equation utilizing the two models will be presented in the following sections.
\subsection{Simulation Setup and Training}
We use a fully-connected four-layer model with a hyperbolic tangent activation function for the approximator/surrogate network. The input layer consists of one neuron (the time coordinate of one collocation point), while the hidden layers comprise $[200,  150, 100, 50]$  neurons, respectively, and the output is the linear combination of output neurons. Although the weights and biases of the networks are initialized randomly at the start of the training, both of them starts from the same initial condition to ensure a fair comparison. As a reminder, the output of the approximator/surrogate network is the approximate solution to our problem. The residual network is a graph encoding the swing equation and source term and provides the loss function (\ref{eq: tot-PINN-loss}) or (\ref{eq: tot_gPINN-loss}) to drive the approximator/surrogate network’s optimization in PINN or gPINN, respectively. The residual points $|\mathcal{T}_f|$ are distributed randomly in the interior of computational domain (time domain of $t \in [0, 20s]$) at which the swing equation should hold and the initial condition is incorporated with $50$ values ($|\mathcal{T}_b|$). In addition to an initial training set, to evaluate the neural networks performance, we also need an extensive test data set. To create the true solutions to the swing equation, the \textit{Scipy.integrate} package in Python is used as an ODE numerical solver. Here, the voltage magnitudes $E_g$ and $E_\infty$ are equal to $1$ p.u. and $B_{g\infty} = 0.2$ p.u. The models were implemented and trained using the Pytorch package in Python $3.7$ on a personal computer (Apple MacBook Pro with M1 chip), the Adam optimizer \cite{KingmaAdam} with the default learning rate $lr = 10^{-3}$ and a maximum of $20000$ epochs in all the cases barring the transfer learning case study. We used the $L^2$ relative error of the prediction of $u$ and $\omega$ and the $L^1$ relative error of $m_g$ and $d_g$ to evaluate the performance of the estimating models (proposed model over the PINN model). In this study, we choose the weights $w_f = w_b = w_i = 1$. Since the performance of the gPINN is sensitive to the choice of the weight $w_g$ for solving the swing equation, we performed the network training using different values of the weight $w_g$, including $1$, $0.1$, and $0.01$. By assessing the relative $L^2$ error between the predicted and the exact solution of $\delta(t)$ and $\omega(t)$, the proper value chosen for the $w_g$ is $0.01$ which achieved the lowest error.

We assume system inertia and damping are known in our first part of the study, and that the system is not in equilibrium. Active power input ($P_m$) and the initial condition ($[\delta(t_0), \omega(t_0)]^T$) is specific to each case study and will be defined later on. For different values of power input and initial conditions, the system may be stable, become unstable, or multiple oscillations may occur. Therefore, it is crucial to achieve high accuracy in each of these regimes. In the second part, inertia and damping are also unknown parameters. Given scattered observed data about angle measurements, our goal is to identify the parameters $m_g$ and $d_g$ of (\ref{eq: gPINNswing}), as well as to obtain the trajectory of $\delta$.
\vspace{-3mm}
\subsection{Forward ODE Problem - Prediction Accuracy in Capturing Power System Dynamics}
\begin{figure}[t]
\centering
\includegraphics[width=.8\linewidth, trim={0.cm 0.5cm 0.5cm 1cm},clip]{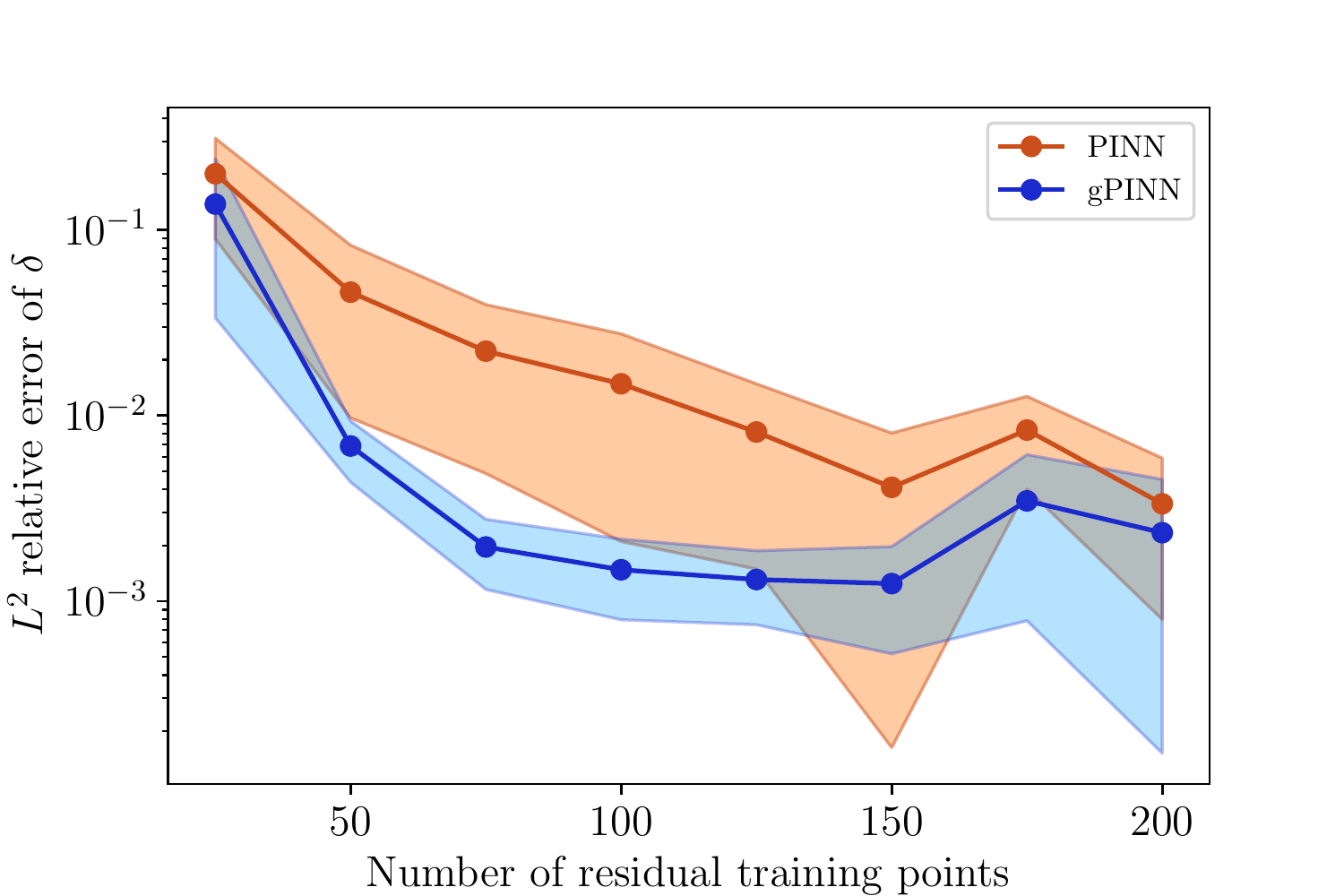}
\includegraphics[width=.8\linewidth, trim={0.cm 0.cm 0.50cm 1cm},clip]{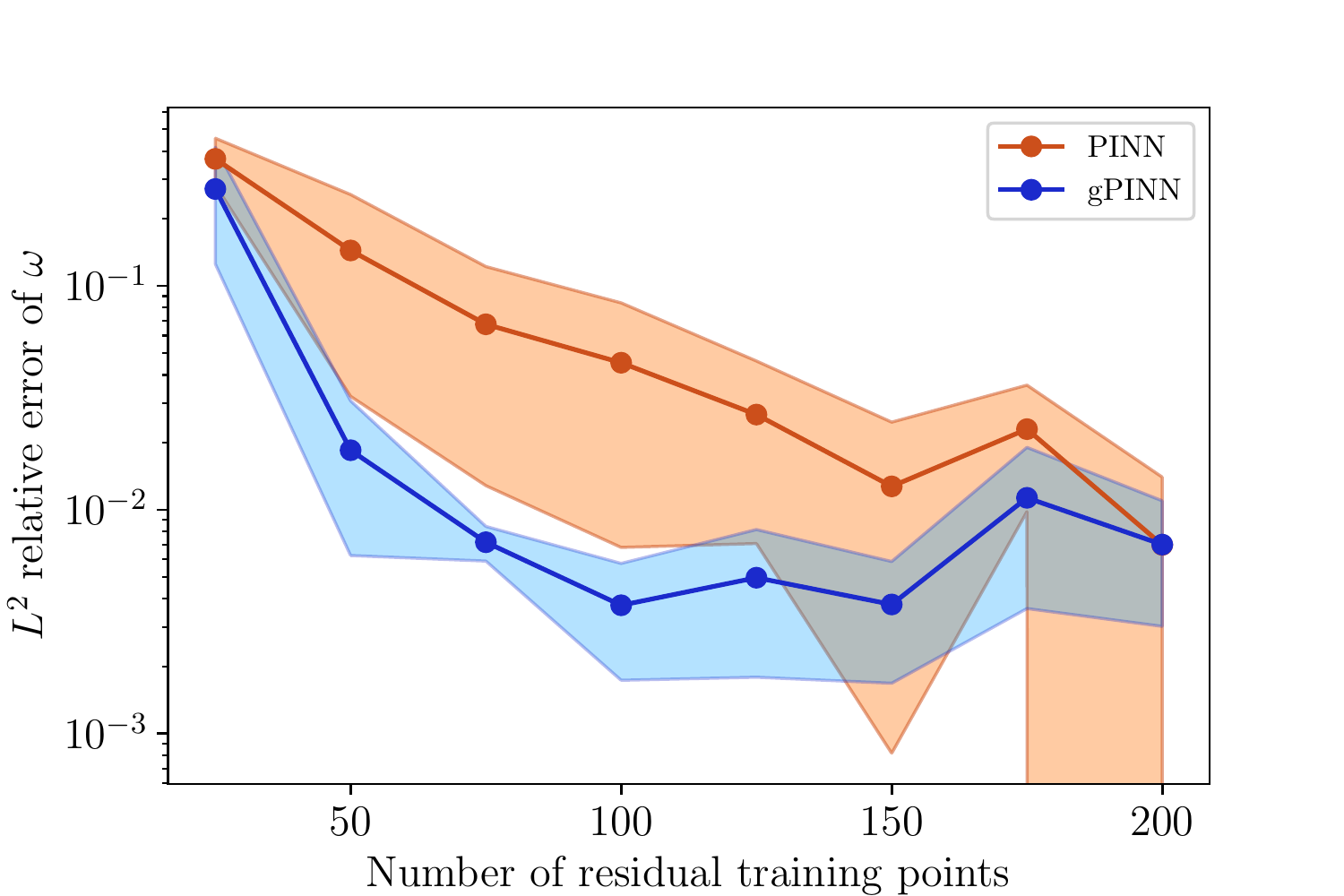}
\caption{$L^2$ relative error of $\delta$ and $\frac{\partial \delta}{\partial t}=\omega$ for the PINN and gPINN. The shaded regions represent one standard deviation across 10 random runs.}
\vspace{-5mm}
\label{fig: err_num_points}
\end{figure}

Figure \ref{fig: err_num_points} depicts the $L^2$ relative error of prediction $\delta(t)$ and $\omega(t)$ for PINN and gPINN when they are trained with different number of residual points. The mean and one standard deviation of 10 independent runs are represented by the line and shaded region. Since increasing the number of residual points in our simulations may lead to over-fitting to the training data, this observation can shed light on choosing the number of appropriate residual points for solving our specific problem that is the swing equation in (\ref{eq: gPINNswing}). Note that this results are obtained for the case that the given generator is in the stable mode of operation. It is shown that when we use more training points up to $150$ points, both gPINN and PINN have smaller $L^2$ relative error of the prediction of $\delta(t)$, and gPINN performs significantly better than PINN at an error of around one order of magnitude smaller. Furthermore, the gPINN outperforms the PINN in terms of predicting the derivative $\frac{d \delta(t)}{d t} = \omega(t)$. The standard deviation in the gPINN has a tighter bound which represents the robustness of the proposed method in predicting the target variables. Finally, we chose $150$ residual points as the base line for training of the models in the rest of the case studies. 

Table \ref{tab: relative error} shows the $L^2$ relative error ($\times 10^2$) for the prediction of $\delta(t)$ and recovered $\omega(t)$ of 10 independent trials, where the performance of the gPINN algorithm over the PINN algorithm is demonstrated in each considered condition: stable, unstable, or oscillating. Note that, to compute the frequency $\omega(t)$, we use the AD to extract the frequency directly from the gPINN or PINN. Hence, the more accurate prediction of the $\delta(t)$ will result in a better estimation of the frequency of the system.  The table reports the comparison of best, average, and worst results that are obtained by these two models over the test set. It is observed that minimum, average, and maximum $L^2$ error for prediction of the $\delta(t)$ offered by the gPINN is lower than corresponding values obtained from the PINN, with the exception of the ``stable'' case study for the best load angle estimation. An even greater improvement by the gPINN can be seen in the $L^2$ relative error of $\omega(t)$. Further, the worst $L^2$ relative error obtained by the gPINN is lower than the average results of the PINN in almost all the cases.
As these value indicate, contrary to the gPINN, the PINN struggles with predicting the $\delta(t)$ with high accuracy and subsequently recovering the frequency $\omega(t)$ when the system is facing an oscillating condition. Most of the time it fails to predict the target variables with an acceptable level of confidence for making and informed decision. The calculated standard deviation of the error for the gPINN is considerably less than the PINN (up to $5$ and $10$ times smaller in the oscillating case for predicting $\delta(t)$ and $\omega(t)$ respectively), which clearly indicates a small variation range for prediction values and robustness of the gPINN. The noted results prove the superiority of the proposed gPINN for solving the general swing equation problem as compared with the base PINN model in different conditions of the systems.
\begin{table*}[ht!]
    \centering \small
    \vspace{2mm}
\caption{Comparison of $L^2$ relative error ($\times 10^{-2}$) for prediction of $\delta$ and recovered $\omega$ of two models based on $10$ trials.}
\label{tab: relative error}
\begin{tabular}{l c c c c c c c c c}
\toprule
\multirow{2}{*}{Model}  & \multirow{2}{*}{Test Case } & \multicolumn{4}{c}{\textbf{$\delta$}} & \multicolumn{4}{c}{\textbf{$\omega$}} \\  \cmidrule(lr{0.5em}){3-6} \cmidrule(lr{0.5em}){7-10} 
        &    &  Min & Average & Max & \makecell{standard \\deviation} &  Min &  Average &  Max & \makecell{standard\\deviation}\\
\midrule
\multirow{3}{*}{gPINN} & Stable & 0.054 & 0.533 & 1.467 & 0.506 & 0.217 & 0.705 & 1.278 & 0.441\\
                       & Unstable & 0.123  & 0.481 & 1.655 & 0.481 & 0.147 & 0.737 & 2.356 &  0.708\\
                       & Oscillating & 0.994 & 1.794 & 4.920 & 1.198 & 2.448 & 4.092 & 7.395 & 3.049\\
\midrule
\multirow{3}{*}{PINN} & Stable & 0.048 & 0.916 & 3.185 & 1.027  & 0.152 & 2.908 & 9.931 & 3.258\\
                       & Unstable & 0.811  & 3.758 & 8.181 & 2.586 & 1.517 & 4.763 & 10.202 &  3.204\\
                       & Oscillating & 3.304 & 22.946 & 38.792 & 16.536 & 7.653 & 54.415 & 94.681 & 39.243\\
\bottomrule
\end{tabular}
\vspace{-3mm}
\end{table*}

\begin{figure}[t]
\centering
  \includegraphics[width=.48\linewidth, trim={0.2cm .8cm 1.15cm 0.88cm},clip]{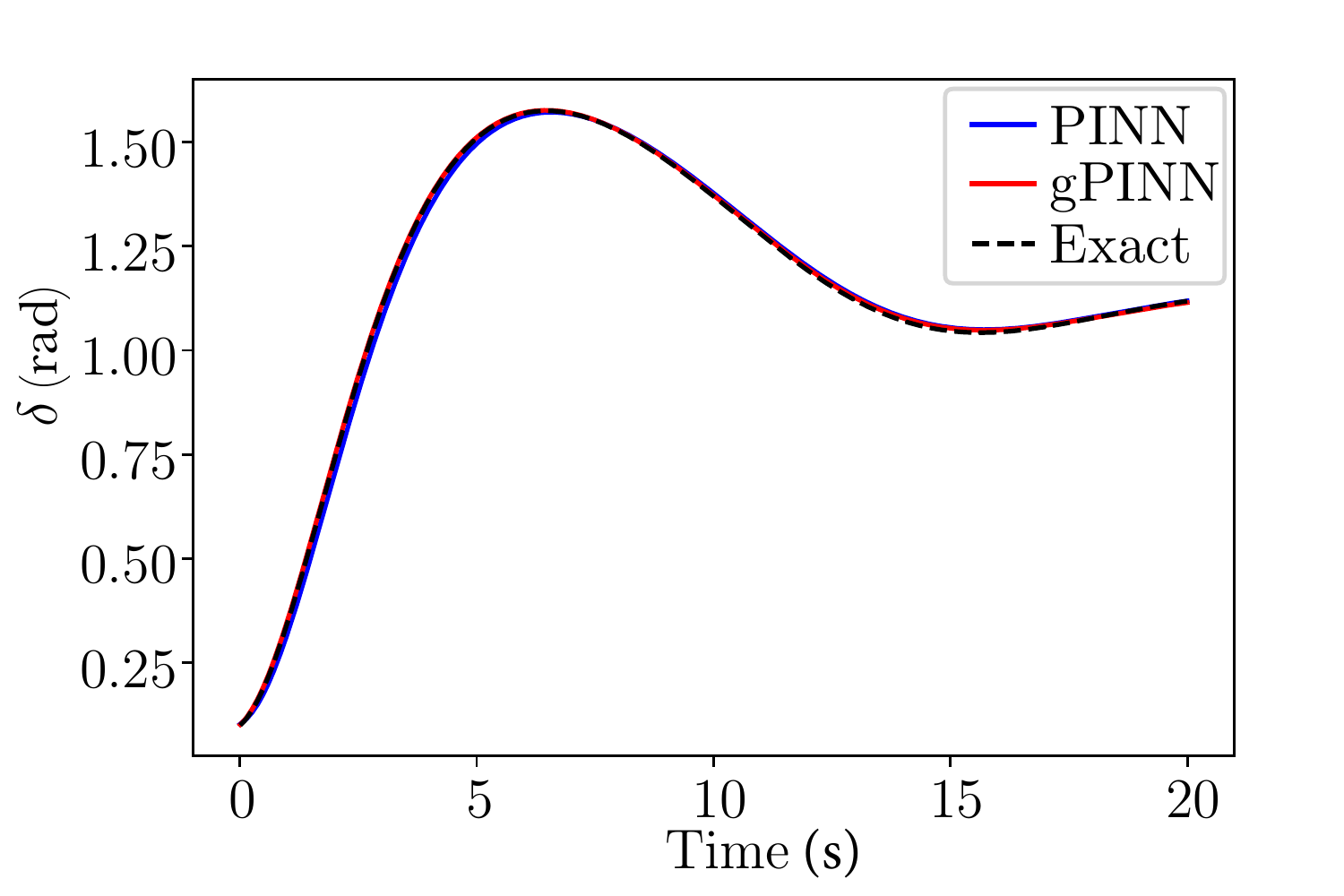}    
  \includegraphics[width=.48\linewidth, trim={0.2cm 0.8cm 1.3cm 0.88cm},clip]{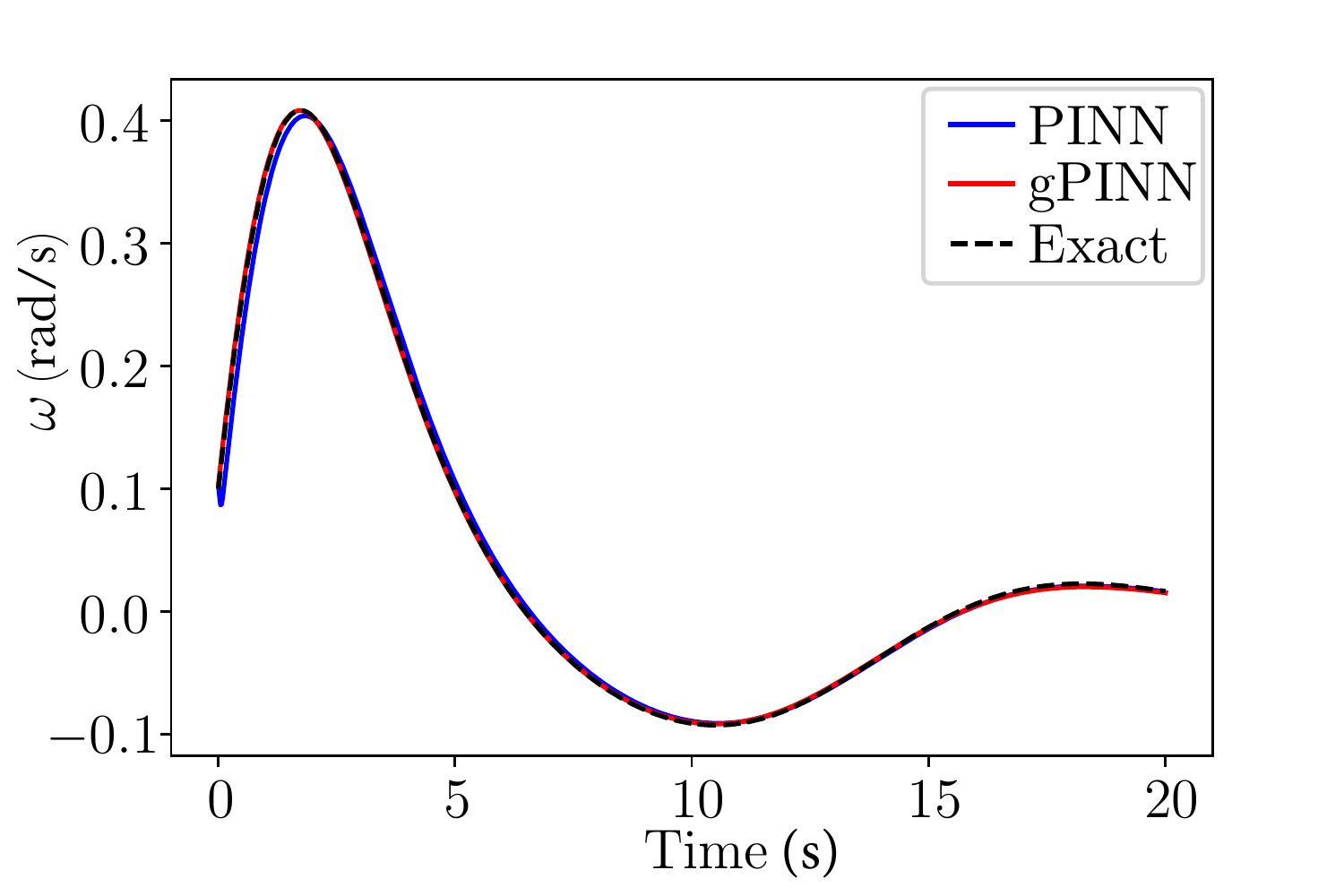} \\
  \includegraphics[width=.48\linewidth, trim={0.2cm .8cm 1.28cm 0.8cm},clip]{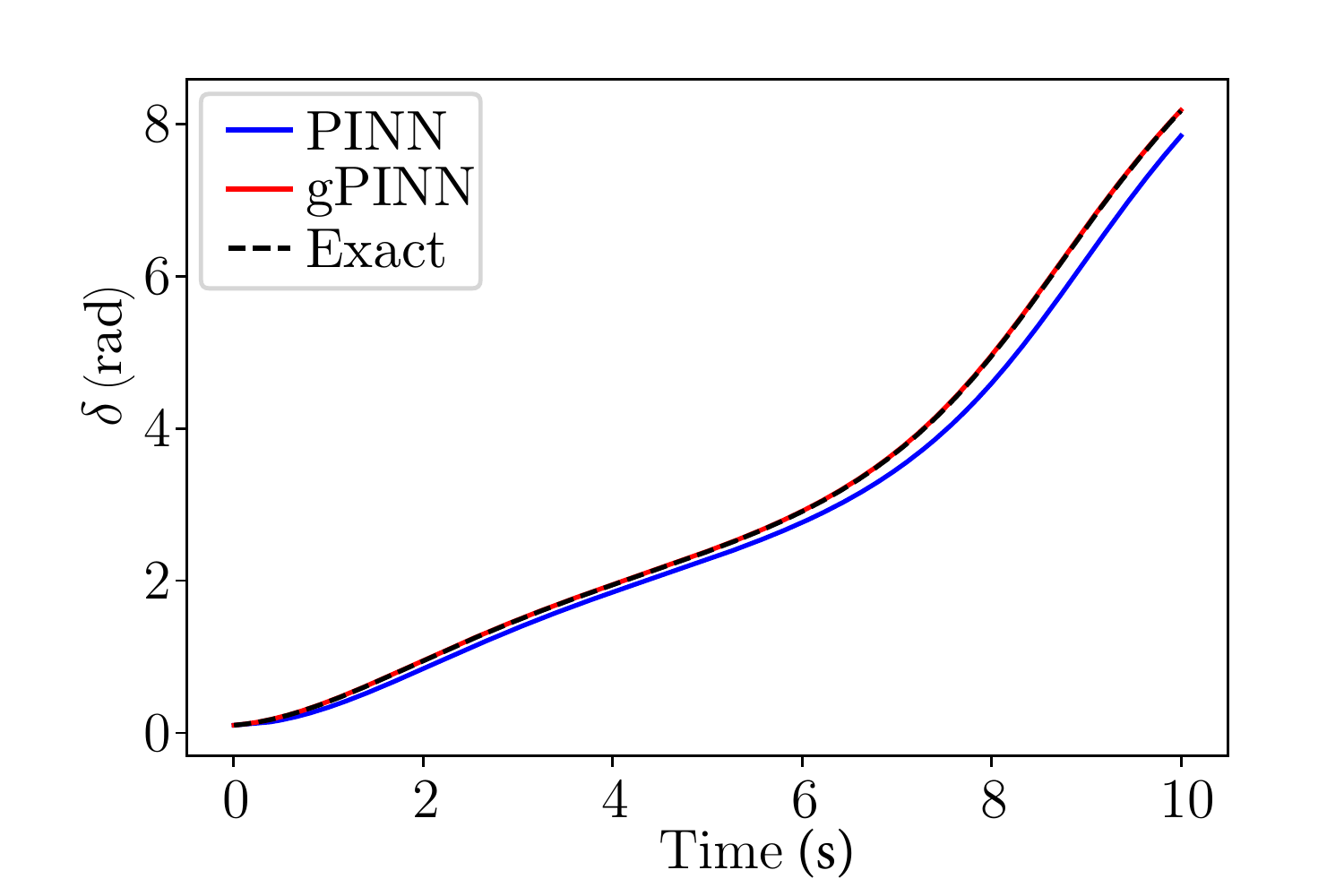}    
  \includegraphics[width=.48\linewidth, trim={0.2cm 0.8cm 1.2cm 0.8cm},clip]{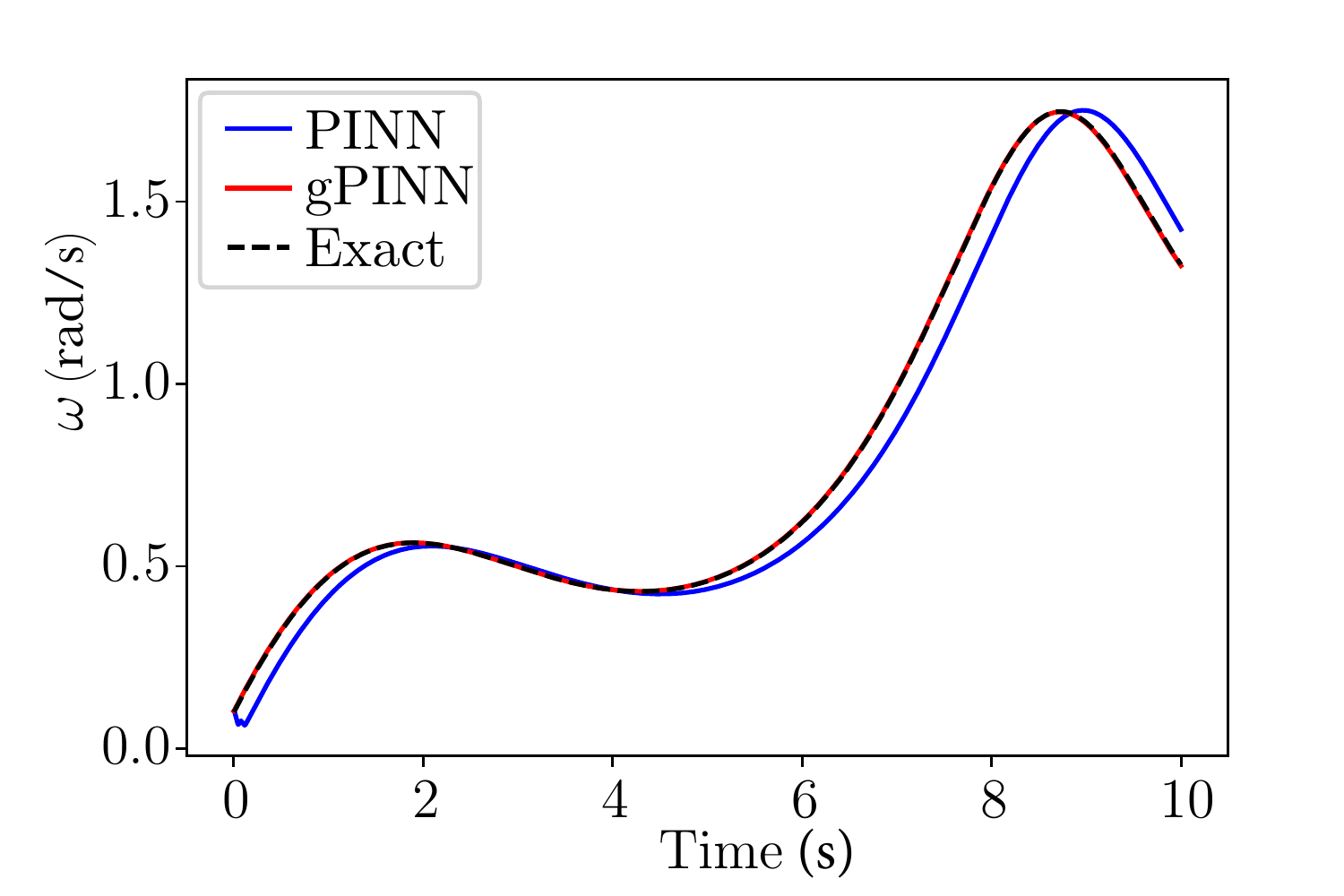}\\
     \includegraphics[width=.48\linewidth, trim={0.2cm .1cm 1.1cm 0.6cm},clip]{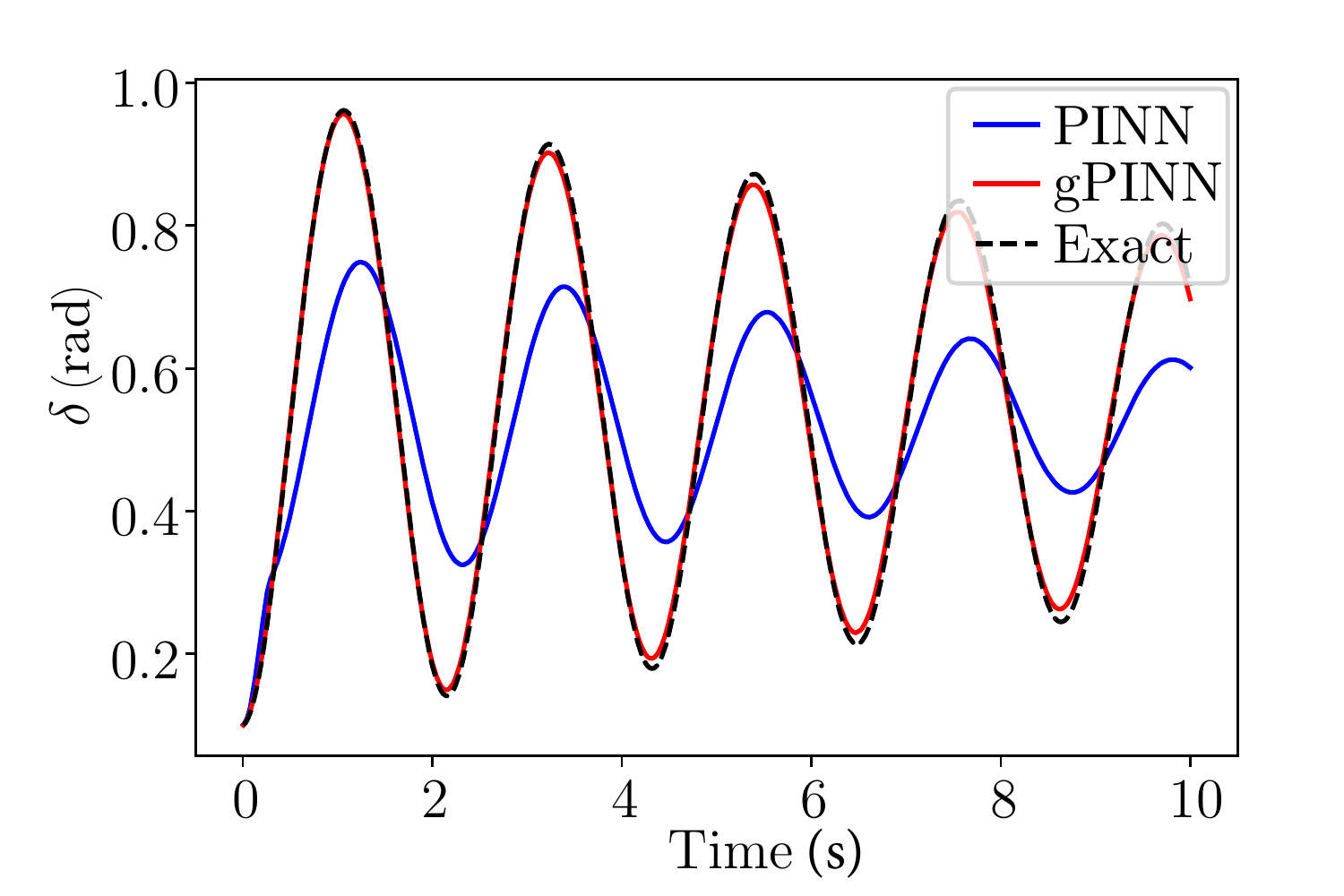}   
  \includegraphics[width=.48\linewidth, trim={0.2cm .1cm 1.1cm 0.6cm},clip]{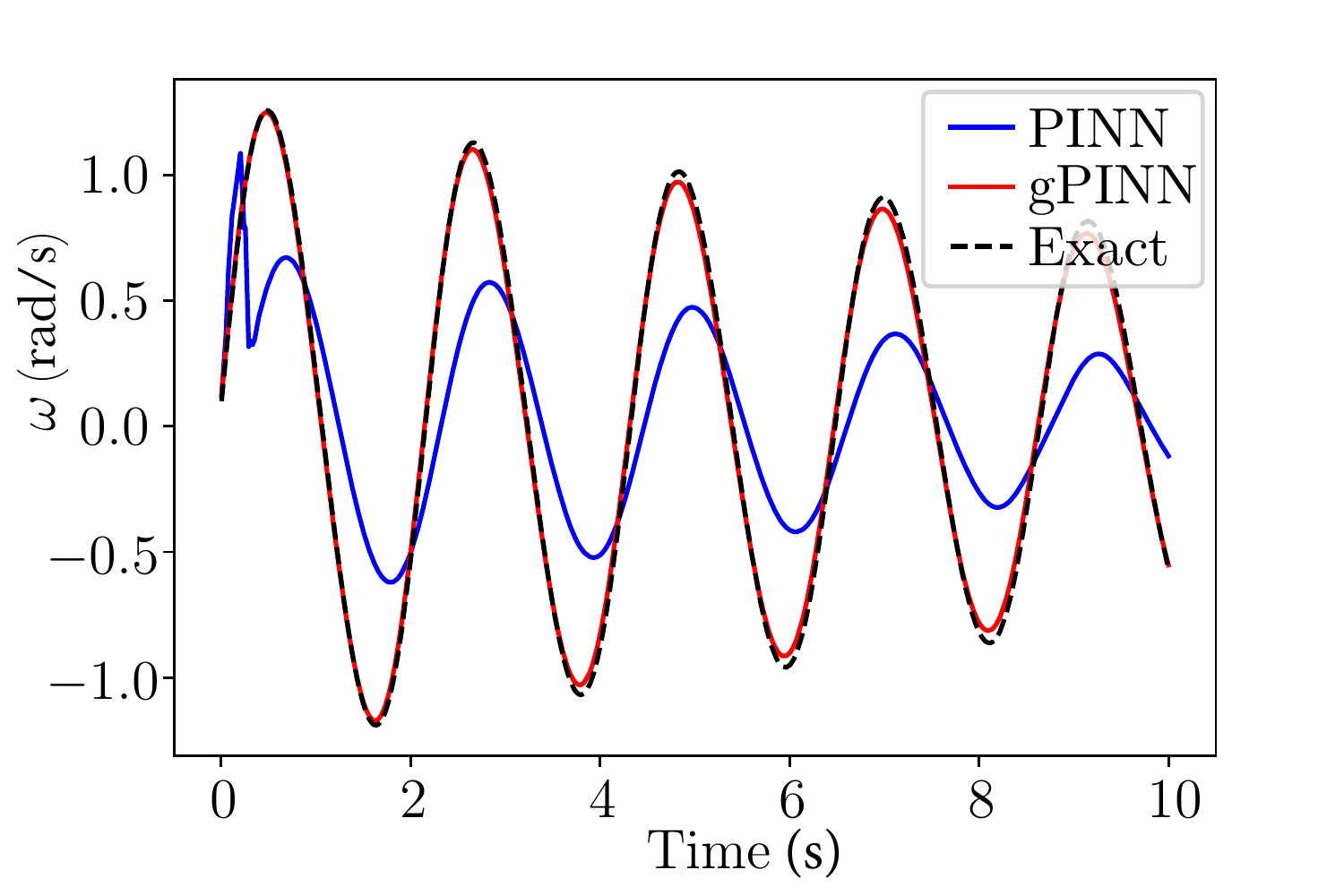} 
\caption{Example of the predicted $\delta(t)$ (left) and corresponding $\omega(t)$ (right) in three cases. As expected, the prediction error increases as the distance into the future increases.}
\vspace{-6mm}
\label{fig: delta_pred}
\end{figure}

Figure \ref{fig: delta_pred} depicts the swing equation's approximate solution with the actual trajectory of the load angle $\delta(t)$ (in the left side) and the corresponding recovered frequency signal $\omega(t)$ (in the right side) for the gPINN and PINN models on the test set.  Figure \ref{fig: delta_pred} (a)-(c) shows the predicted load angle and frequency for the three possible states of the generator as mentioned in Table \ref{tab: relative error}. In all these cases, the trajectory starts from the same initial condition $[\delta(t_0), \omega(t_0)]^T = [0.1, 0.1]^T$ and only the input mechanical power is different which puts the system in different condition. Moreover, the depicted approximate solutions have the $L^2$ relative error close to the average value reported in the Table \ref{tab: relative error} to provide a pictorial overview of the ability of models in estimating the desired variables of interest. In the left figures, we show the trajectory of the $\delta(t)$, with a $L^2$ relative error of $0.94\cdot10^{-2}$ and $0.23\cdot10^{-2}$ in stable, $5.89\cdot10^{-2}$ and $0.19\cdot10^{-2}$ in unstable,  $25.35\cdot10^{-2}$ and $1.87\cdot10^{-2}$ in oscillation state for both PINN and gPINNs, respectively. 

It can be found that the gradient-enhanced physics-informed neural network is able to predict the trajectory of the angle $\delta(t)$ with high accuracy, and that the frequency signal $\omega(t)$ can be successfully recovered using automatic differentiation in all three examples. Interestingly, when we analyze the accuracy of the prediction $\delta(t)$ or the recovered $\omega(t)$ in all the three aforementioned conditions, we see that the PINN tends to underestimate the peaks and troughs (specifically in the unstable or oscillating state in \ref{fig: delta_pred} (b) and (c)). Intuitively, these predictions are worsened as we make predictions further into the future. Overall, gPINN model outperforms the PINN in terms of both predicting $\delta(t)$ and obtaining corresponding $\omega(t)$, since gPINN utilizes the information of the gradient and thus has a much faster convergence rate than PINN. In other words, the gPINN also has computational benefits versus the PINN. Therefore, using only a handful of initial data, the gradient-enhanced physics-informed neural network can accurately capture the intricate nonlinear behavior of the swing equation equation.
 

\subsection{The Importance of Transfer Learning}
\begin{figure}[t]
\centering
\includegraphics[width=.8\linewidth, trim={0.cm 0.cm 0.5cm 1cm},clip]{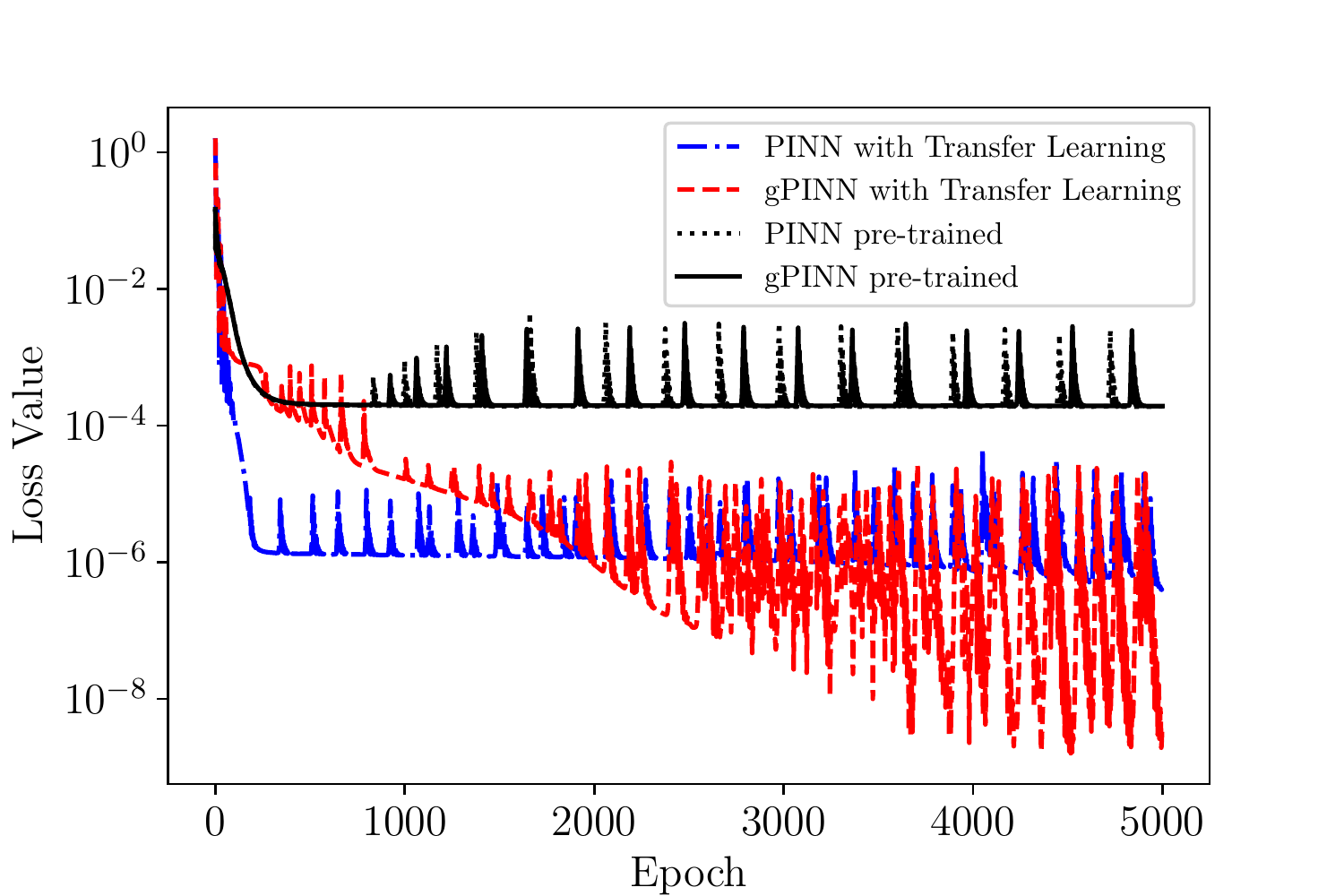}
\caption{Training error with transfer learning for the gPINN/PINN models and their pre-trained test cases.}
\vspace{-3mm}
\label{fig: transfer_loss}
\end{figure}

\begin{figure}[t]
\centering
\includegraphics[width=.8\linewidth, trim={0.2cm 0.5cm 0.5cm 1cm},clip]{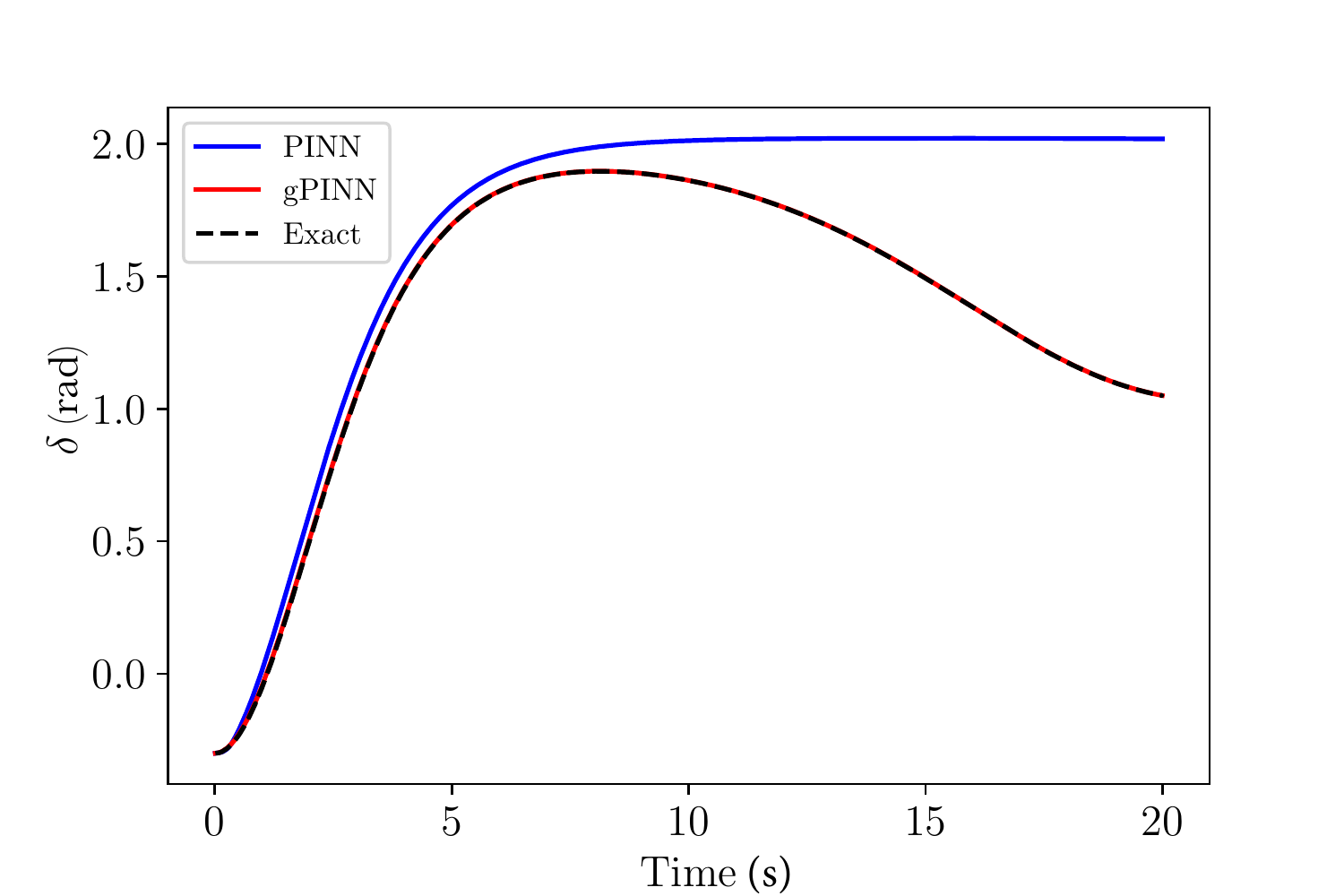}
\includegraphics[width=.8\linewidth, trim={0.2cm 0.0cm 0.50cm 1cm},clip]{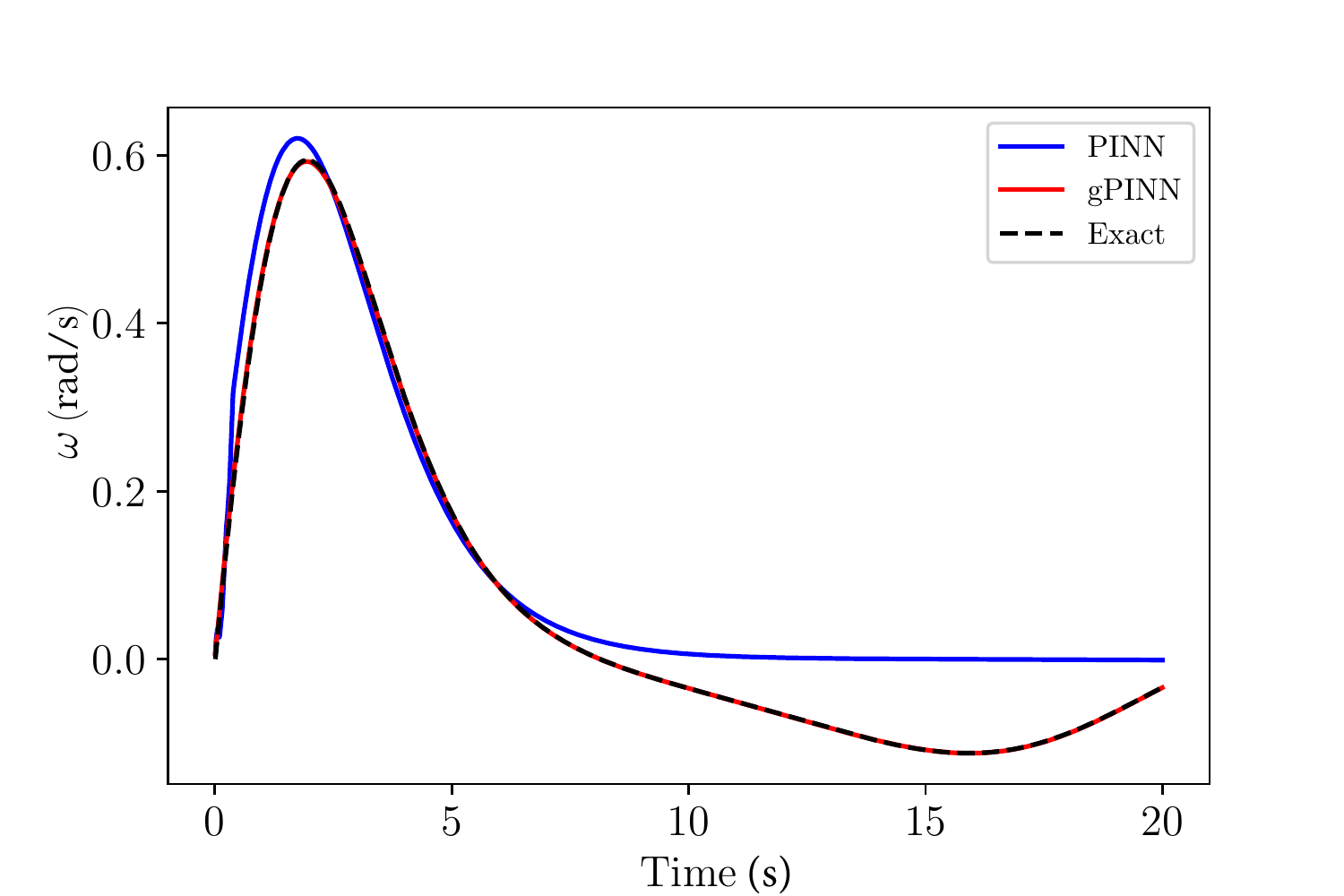}
\caption{The predicted $\delta$ and $\omega$ from PINN and gPINN after $5000$ training epochs using transfer learning.}
\vspace{-6mm}
\label{fig: transfer_pred}
\end{figure}

Next, we discovered that the utilization of transfer learning technique is critical to speed up training and improve the performance of our proposed model with a reduced number of epochs. The transfer learning method entails using pre-trained models as the starting point on specific tasks (e.g. solving a swing equation) given the vast computational and time resources required to develop neural network models on these problems and from the huge jumps in skill that they provide on related problems. The first completely trained network weights and biases, which is trained for predicting $\delta(t)$, can then be used to initialize the gPINN/PINN network we want to solve. In this way, the first gPINN/PINN passes on the encoding knowledge it has gained to the second gPINN/PINN. To demonstrate the benefit of transfer learning in gPINN/PINN, we solve a test case with a source term in a stable state. We initialize the networks, corresponding to the model itself, with the results obtained during the training with Eq. (\ref{eq: gPINNswing}) as a source term. When not using transfer learning, we have trained the gPINN/PINN with $20$,$000$ epochs of the Adam optimizer to have acceptable performance like the state of the art studies in this domain. But when using a transfer learning technique each model is only trained over $5$,$000$ epochs which therefore indicate a quicker training. Figure \ref{fig: transfer_loss} depicts  a comparison of the training error for the gPINN and PINN network initialized with their corresponding pre-trained models (in their first $5$,$000$ epochs), e.g., without transfer learning and with transfer learning. In this case, transfer learning usage allows gaining at least two orders of magnitude improvement in the training error in less than $1$,$000$ epochs. Further, it is found that utilization of the transfer learning leads to an initial super-convergence to a relatively low training error. As a result, transfer learning is a crucial operation for gPINN/PINN to compete with other scientific computing solvers. The performance and convergence behavior of the proposed model suggest that the transfer learning method is a good match for the network learning this problem. Fig. \ref{fig: transfer_pred} illustrates actual trajectories of the $\delta(t)$ and the recovered frequency signal $\omega(t)$, and the approximations learned by the PINN and gPINN  on the test set. It is shown that how much that each model has the ability in tacking the true solution after just $5$,$000$ epochs.  In this case, the relative $L^2$ between the exact and predicted solutions in $\delta(t)$ are $29.3\cdot10^{-2}$ and $0.17\cdot10^{-2}$ for PINN and gPINN models, respectively. These values for the frequency signal $\omega(t)$ are $27.7\cdot10^{-2}$ and $0.75\cdot10^{-2}$. This plot shows that the gPINN has a easy prediction task in finding the true solution compare to the PINN network, indicating that proposed model may effectively capture such behavior, as indeed confirmed in the corresponding low gPINN prediction errors.

\subsection{Inverse Problem of ODE - Discovery of Inertia and Damping Coefficients}
\begin{figure}[t!]
\begin{minipage}[b]{0.48\linewidth}
\centering
\begin{tabular}{l}
  \includegraphics[width=\textwidth, trim={0.2cm .0cm 1.05cm 0.88cm},clip]{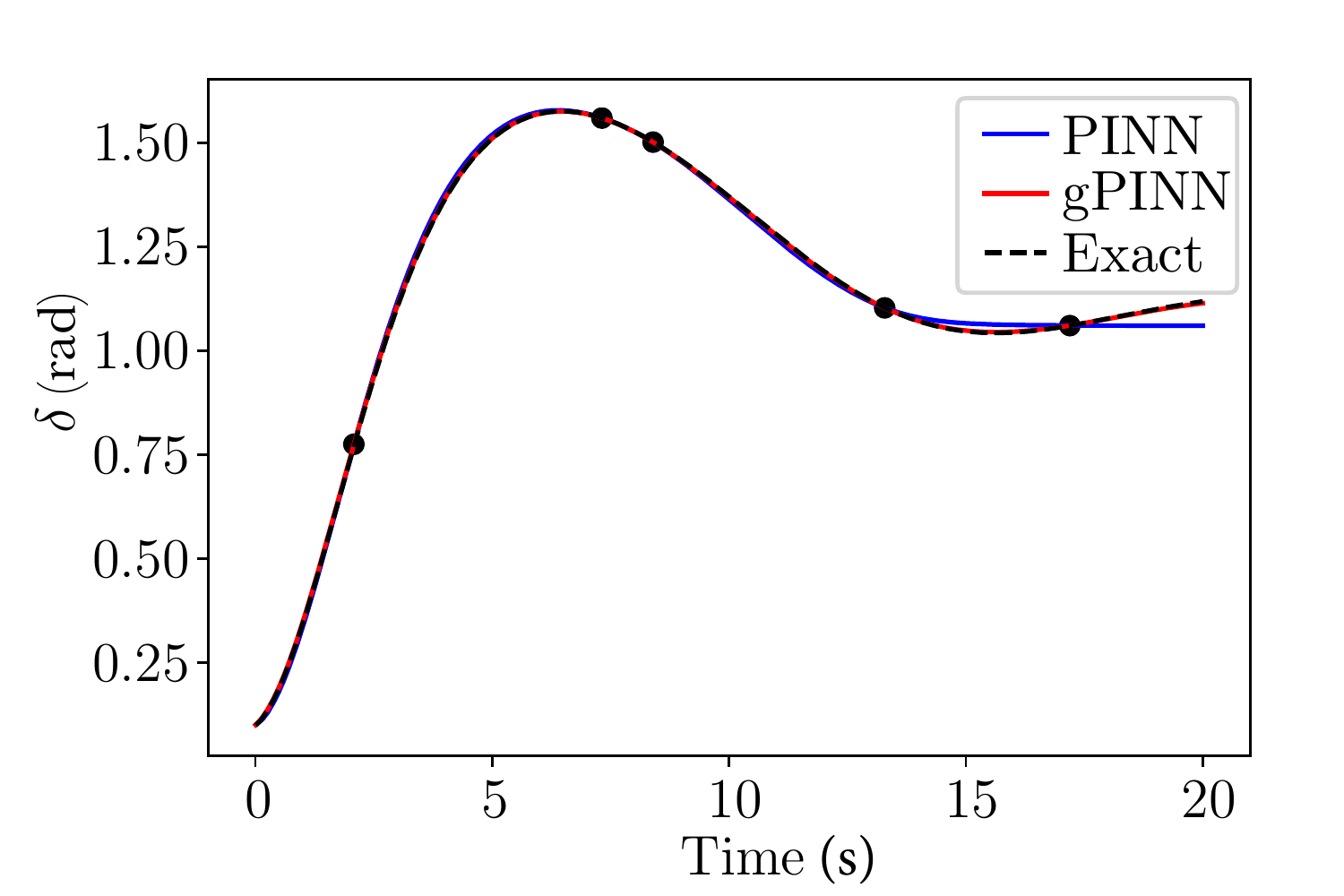}\\
(a) \makecell{\small The predicted $\delta$ \\\small from PINN and gPINN}
\end{tabular}
\end{minipage}
\begin{minipage}[b]{0.48\linewidth}
\centering
\begin{tabular}{l}
  \includegraphics[width=\textwidth, trim={0.2cm .0cm 1.05cm 0.95cm},clip]{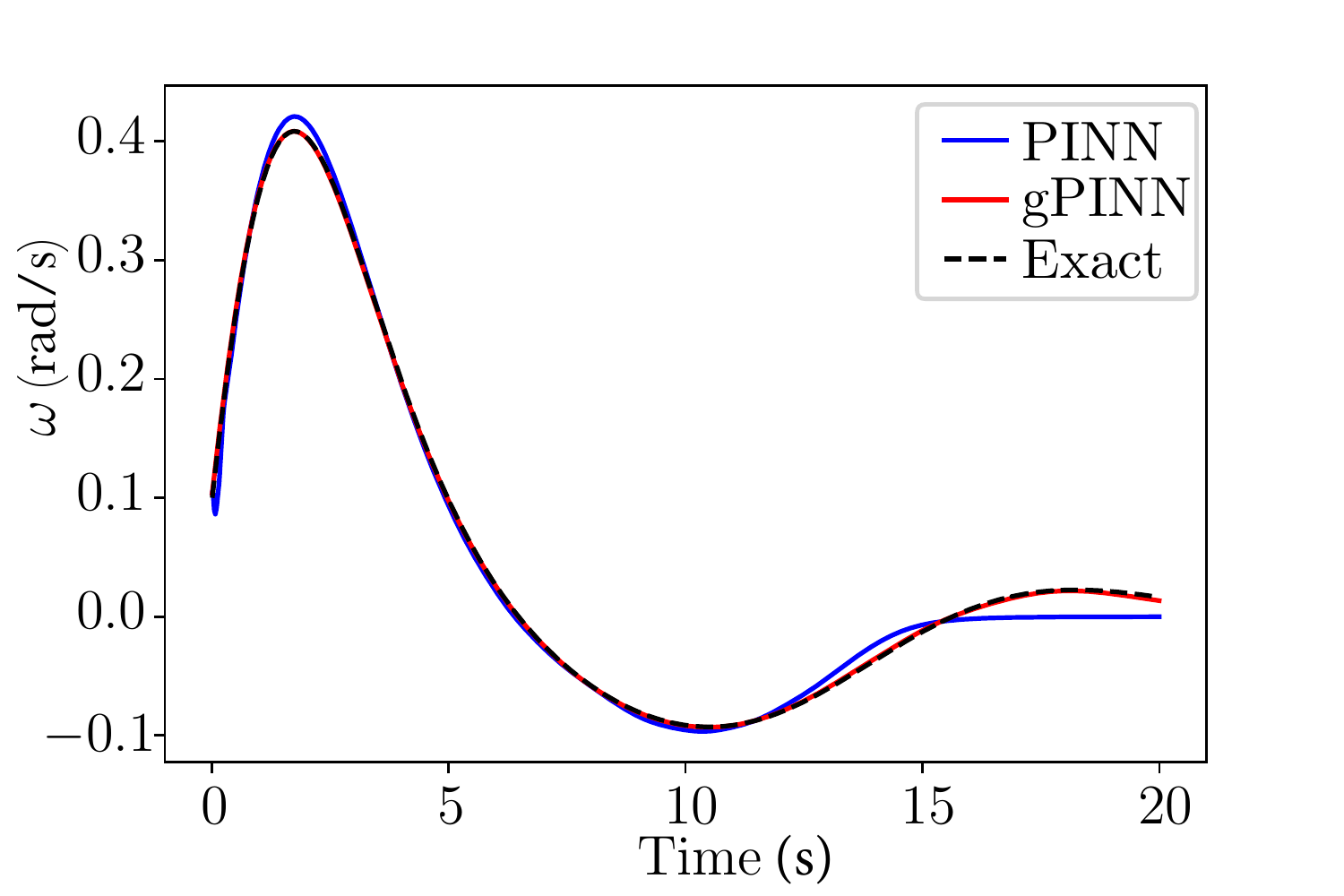}\\
(b) \makecell{\small The predicted $\omega$ \\\small from PINN and gPINN}
\end{tabular}
\end{minipage}
\newline
\begin{minipage}[b]{0.48\linewidth}
\centering
\begin{tabular}{l}
  \includegraphics[width=\textwidth, trim={0.2cm .0cm 1.1cm 0.88cm},clip]{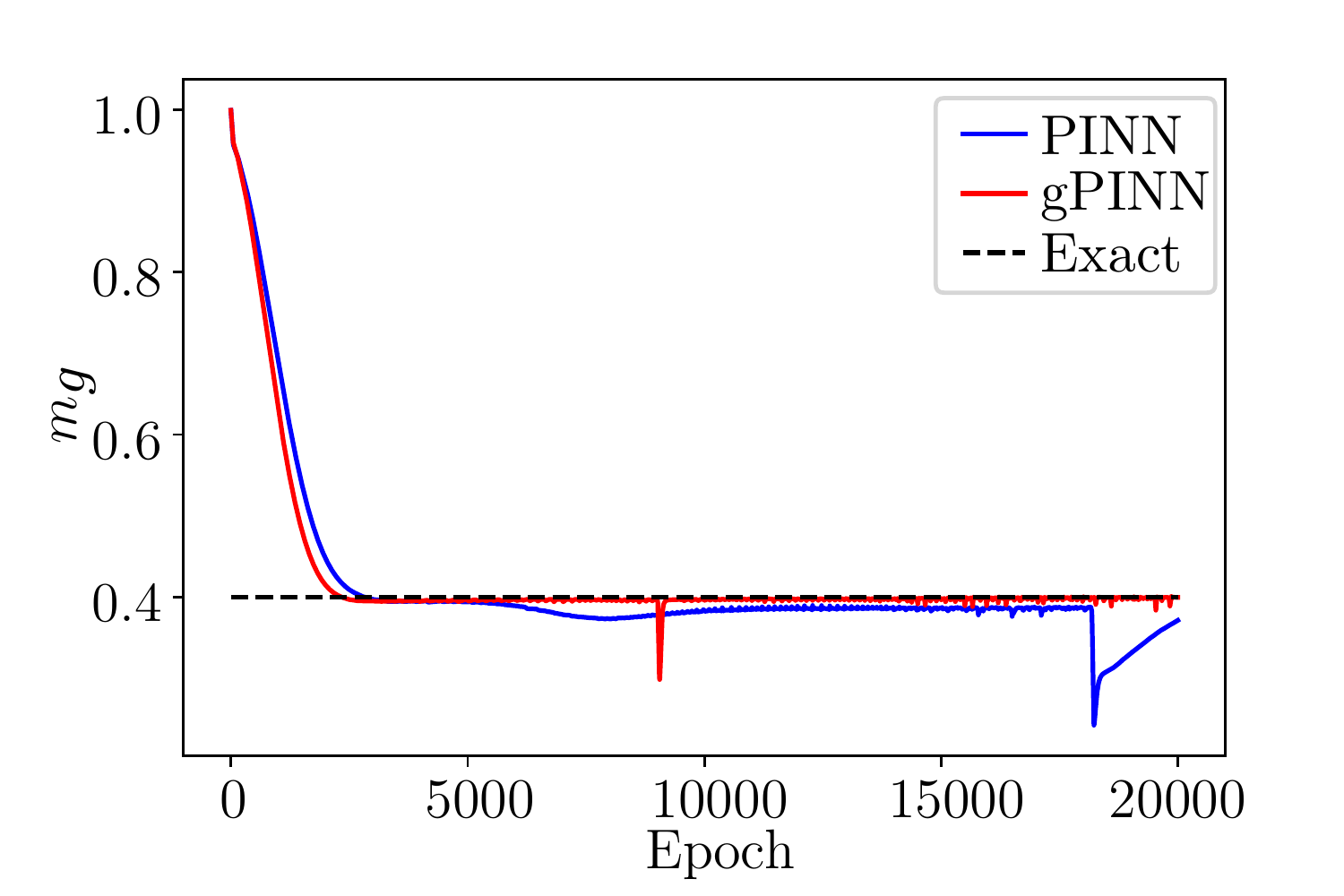}\\
(c) \makecell{\small The convergence of the \\\small predicted value for $m_g$}
\end{tabular}
\end{minipage}
\begin{minipage}[b]{0.48\linewidth}
\centering
\begin{tabular}{l}
  \includegraphics[width=\textwidth, trim={0.25cm .0cm 1.0cm 0.88cm},clip]{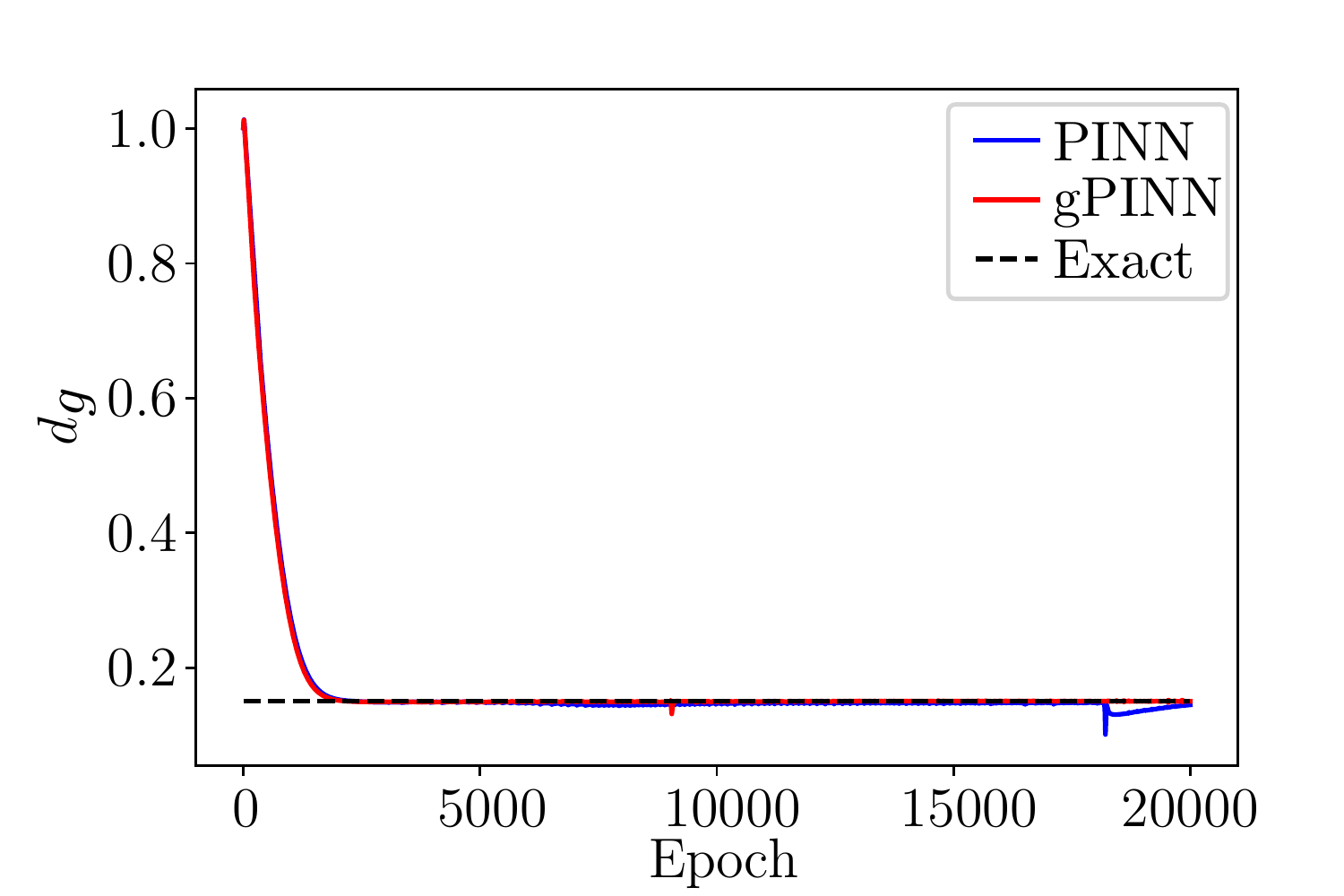}\\
(d) \makecell{\small The convergence of the \\\small predicted value for $d_g$}
\end{tabular}
\end{minipage}
\caption{Inferring both $m_g$ and $d_g$ throughout training. The black dots in (a) show the observed locations of $\delta$.}
\label{fig: iverse-pred}
\end{figure}

Finally, the performance in predicting system inertia and damping from observed trajectories is evaluated. Given scattered and potentially noisy data on the $\delta(t)$ component, our goal is to identify unknown parameters $m_g$ and $d_g$, as well as to obtain a qualitatively accurate reconstruction of $\delta(t)$. To infer $m_g$ and $d_g$, we have chosen the data measurements of the load angle $\delta(t)$ in only $5$ time steps, which are randomly selected in the simulation horizon to highlight the ability of our method to learn from scattered and scarce training data. The neural network architectures used here are the same as the cases for forward problem. A visual comparison against the exact load angle and frequency solutions is presented in Fig. \ref{fig: iverse-pred}. Similar to what we observed in the forward swing equation problem, the gPINN outperforms the PINN in this case. While the PINN failed to predict $\delta(t)$ and $\omega(t)$ near the peaks and end of the simulation, the gPINN has decent accuracy. Moreover, it is shown that during training, the predicted $m_g$ and $d_g$ in PINN did not converge to the true value (in the case of predicting $d_g$, the PINN performs better), while with the gPINN, the predicted $m_g$ and $d_g$ of a system is more accurate. Note that the predicted parameters will have an effect on the relative error of the predicted variables $\delta(t)$ and the recovered frequency signal $\omega(t)$. For a better illustration and test the robustness of the models, Table \ref{tab: inverse-param} compares the performance of the gPINN and PINN algorithms in terms of minimum, average, maximum and standard deviation of $L^1$ relative error of the predicted parameters $m_g$ and $d_g$ based on 10 independent trials for the stable state of the system. It is found that the best $L^1$ relative error in predicting both parameters is offered by the gPINN algorithm ($0.000\cdot10^{-2}$ and $0.008\cdot10^{-2}$ for $m_g$ and $d_g$, respectively). It is observed that minimum, average, and maximum $L^1$ relative error offered by the gPINN algorithm is lower than corresponding values obtained from PINN algorithm.
This finding of inferring a continuous quantity of interest from auxiliary measurements by exploiting the underlying physics demonstrates promise in handling high-dimensional inverse problems.

\begin{table}[t!]
    \centering
    \vspace{-2mm}
\caption{Comparison of $L^1$ relative error ($\times 10^{-2}$) for prediction of moment of inertia $m_g$ and damping coefficient $d_g$ of two models based on $10$ trials.}
\label{tab: relative error_inverse}
\resizebox{\columnwidth}{!}{\begin{tabular}{l c c c c c}
\toprule
Model & Parameter &  Min & Average & Max & \makecell{standard \\deviation} \\
\midrule
\multirow{2}{*}{gPINN} & $m_g$ & 0.000 & 0.181 & 0.615 & 0.222\\
                       & $d_g$ & 0.008  & 0.195 & 0.456 & 0.167\\
\midrule
\multirow{2}{*}{PINN} & $m_g$ & 0.061 & 1.432 & 7.150 & 2.124 \\
                       & $d_g$ & 0.013  & 0.575 & 3.374 & 1.011\\
\bottomrule
\end{tabular}}
\vspace{-3mm}
\label{tab: inverse-param}
\end{table}

\section{Conclusion}\label{sec: conclude}
In this paper, we presented a framework based on the gradient-enhanced physics-informed neural networks (gPINNs) for power system applications specifically in dynamics studies. We demonstrated the effectiveness of gPINN in determining the rotor angle and frequency for a single-machine infinite-bus system in the forward problem case. Our numerical findings from all of the cases show that the considered gPINN outperforms the considered PINN with the same number of training points. Due to the incorporation of the underlying swing equation, gradient-enhanced physics informed neural networks use significantly less training data  while obtaining great accuracy. We also depicted that leveraging transfer learning could effectively optimize the model with a reduced number of epochs necessary for training.

Moreover, one of the key advantages of gPINNs is that they can solve inverse PDE/ODE issues as readily as forward problems in terms of implementation. The results shown successful identification of uncertain system parameters such as inertia and damping from a very limited set of input data.
Our findings suggest that these methods have the potential to be successfully applied in larger systems, opening up a slew of new possibilities for power system security and optimization while maintaining high computational speed and accuracy.
As a future extension of our work, we will move towards physics-aware learning models for Optimal Power Flow (OPF) which is an essential component to make ensure that the optimal solutions provided by machine learning models adhere to practical dynamical constraints.

\bibliography{main_v2.bib}{}
\bibliographystyle{IEEEtran}

\end{document}